\documentclass[letterpaper,10pt,conference]{templates/ieeeconf}

\IEEEoverridecommandlockouts

\usepackage{amssymb} 

\usepackage{graphicx} 
\usepackage{tabularx} 
\usepackage{balance} 
\usepackage[table]{xcolor} 
\usepackage{arydshln} 
\usepackage{cite} 

\newcolumntype{C}{>{\centering\arraybackslash}X}
\newcolumntype{L}{>{\raggedright\arraybackslash}X}
\newcolumntype{R}{>{\raggedleft\arraybackslash}X}

\title{\LARGE\bf
   Haptic-Informed ACT with a Soft Gripper and Recovery-Informed Training for Pseudo Oocyte Manipulation
}

\author{
    Pedro Miguel Uriguen Eljuri$^{1}$,
    Hironobu Shibata$^{2}$,
    Maeyama Katsuyoshi$^{2}$,
    Yuanyuan Jia$^{1}$,\\
    and Tadahiro Taniguchi$^{1, 2}$
    \thanks{
        $^{1}$Pedro Miguel Uriguen Eljuri, Yuanyuan Jia, and Tadahiro Taniguchi are with Kyoto University;
        Yoshida-Honmachi, Sakyou-Ku, Kyoto, Japan. 
        {\tt\small\{urigueneljuri.pedro.5d, jia.yuanyuan.5t\}@kyoto-u.ac.jp, taniguchi@i.kyoto-u.ac.jp} 
    }
    \thanks{
        $^{2}$Hironobu Shibata, Maeyama Katsuyoshi, and Tadahiro Taniguchi are with Ritsumeikan University;
        1-1-1 Noji-Higashi, Kusatsu, Shiga 525-8577, Japan. 
        {\tt\small\{shibata.hironobu,maeyama.katsuyoshi, taniguchi\}@em.ci.ritsumei.ac.jp}
    }
}

\begin{document}


\maketitle
\thispagestyle{empty}
\pagestyle{empty}

\begin{abstract}

In this paper, we introduce Haptic-Informed ACT, an advanced robotic system for pseudo oocyte manipulation, integrating multimodal information and Action Chunking with Transformers (ACT).
Traditional automation methods for oocyte transfer rely heavily on visual perception, often requiring human supervision due to biological variability and environmental disturbances.
Haptic-Informed ACT enhances ACT by incorporating haptic feedback, enabling real-time grasp failure detection and adaptive correction.
Additionally, we introduce a 3D-printed TPU soft gripper to facilitate delicate manipulations.
Experimental results demonstrate that Haptic-Informed ACT improves the task success rate, robustness, and adaptability compared to conventional ACT, particularly in dynamic environments.
These findings highlight the potential of multimodal learning in robotics for biomedical automation.
\end{abstract}

\section{Introduction}
\label{sec:introduction}

Manipulation of cells is the basis for many applications in biological and biomedical engineering.
In cell manipulation, it is required to manipulate and extract a single cell from a cell population for being studied or experimented later~\cite{shakoor2022advanced}.
Different methods are used to manipulate the cells, which can be divided into contact and non-contact methods, such as cell transportation with optical tweezers, micro-fluid technology, or magnetic tweezers~\cite{li2013dynamic,pan2020automated,wang2017three}.
These methods put emphasis on precision and carefulness to avoid damaging or deforming the cells.
An example of cell manipulation in biomedical and biological tasks is oocyte manipulation, where an oocyte is picked from a group to be later used.
Some examples of oocyte manipulations can be found in stem cell research, reproductive medicine (e.g. in vitro fertilization), or regenerative medicine studies~\cite{zhu2017study, zhang2016robotic}.
The related works in automation of oocyte manipulation focus on visual solutions or specialized hardware~\cite{miao2021development,zhu2017study,Liang_Icra_2024} to execute the task under a microscope.
However, human experts are still required to operate or supervise the robot.
Automating the transfer of oocytes using robots offers significant benefits, such as reducing human labor, increasing efficiency, lowering costs, and improving accuracy. 

To achieve this task with a robot, we propose that an expert teaches by demonstration instead of using rule-based methods.
Therefore, we consider it suitable to use imitation learning (IL) to achieve this manipulation task.
However, while IL shows potential in various applications, it still faces significant challenges.
One of the main issues is compounding errors, where small deviations accumulate over time, gradually leading the robot into unfamiliar states not encountered during training, making recovery increasingly difficult once it departs from known trajectories.
In the context of oocyte transfer, these challenges are further exacerbated by the natural variability and unpredictability of biological tasks.
Oocytes' color, size, and position can vary significantly within a cultivation plate or petri dish sample.
Additionally, natural changes in the experimental environment, such as lighting variations and minor disturbances, pose further difficulties.
\begin{figure}[t]
    \centering
    \includegraphics[width=0.99\linewidth]{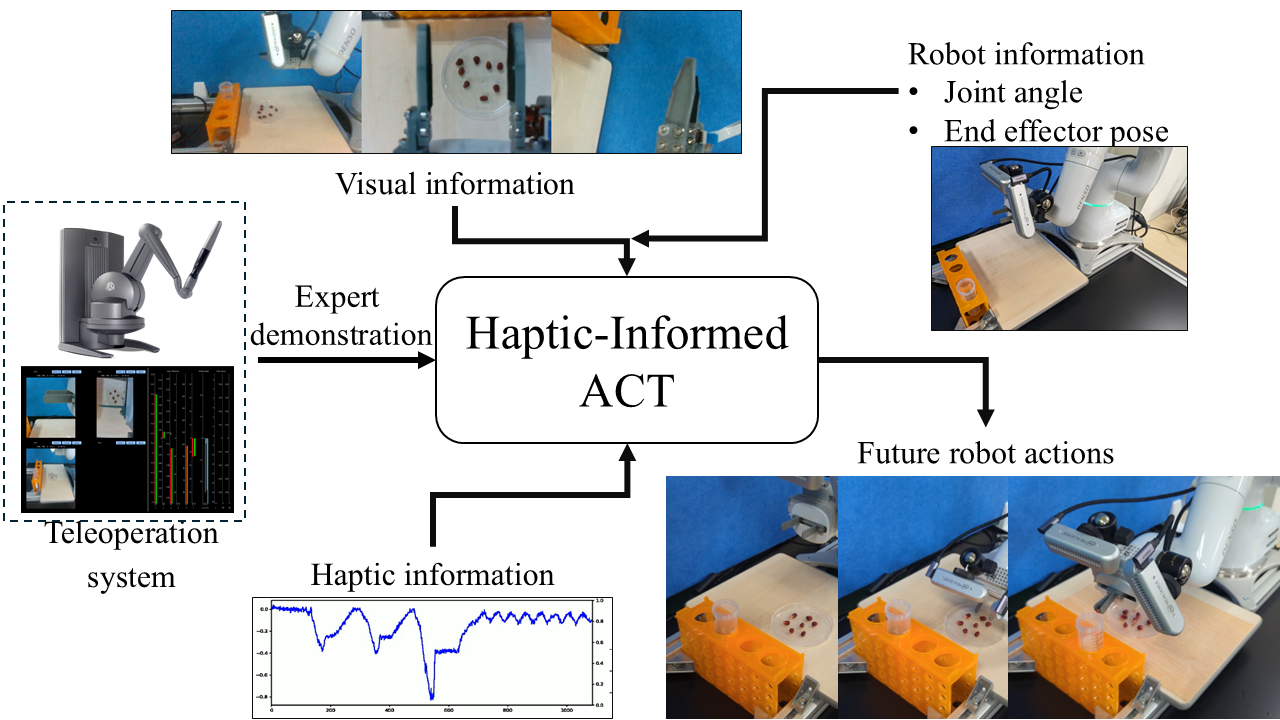}
    \caption{Overview of the proposed Haptic-Informed ACT. It learns from expert demonstration and uses visual, robot, and haptic information to predict the future sequence of robot actions to execute a pseudo oocyte manipulation task.}
    \label{fig:overview}
\end{figure}
\begin{figure*}[t]
    \centering
   \includegraphics[width=0.90\linewidth]{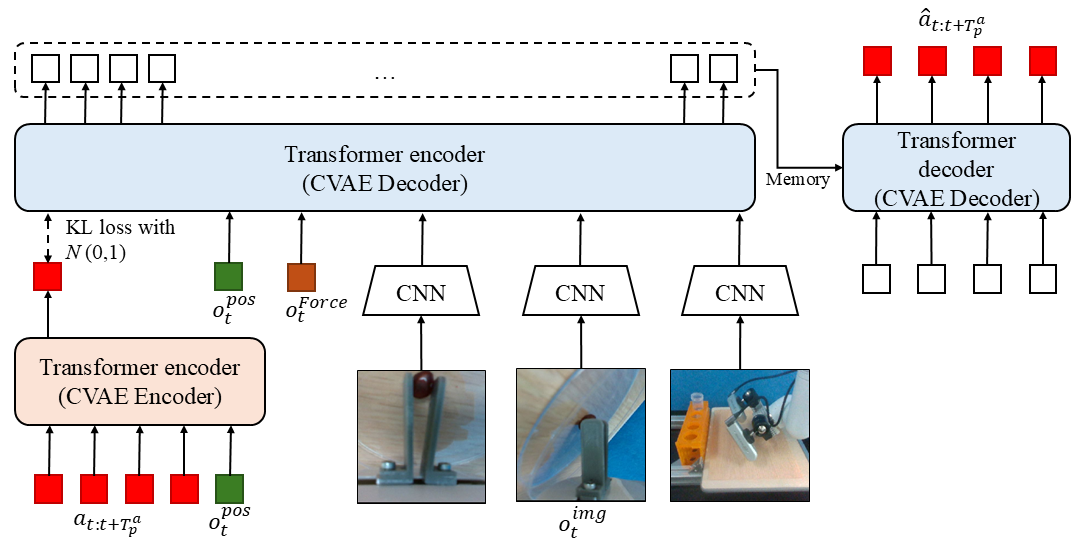}
    \caption{Proposed architecture of Haptic-Informed ACT.}
    \label{fig:proposed_method}
\end{figure*}
We proposed a method that learns from human demonstration using visual and haptic information to overcome these challenges.
Inspired by Action Chunking by Transformers (ACT)~\cite{zhao_learning_2023}, a model proposed by Zhao~\textit{et~al.} 
Compared to other models ACT requires a low amount of samples to learn how to execute a task.
ACT predicts the target joint position or action sequence for the next $k$ time steps instead of just predicting just one step.
However, using only visual and robot information is not enough to achieve this task, specifically during the picking of the oocyte. 
We propose to solve this by introducing haptic information together with visual and robot information.
Furthermore, we propose to use successful expert demonstrations together with samples of cases when the expert fails and needs to execute a recovery maneuver to pick the oocyte again.
Finally, to be able to handle the delicate materials, we introduce a soft gripper that adapts to the variations of size and shape of the target.

The overview of our proposed system is shown in Fig.~\ref{fig:overview}.
Haptic-Informed ACT learns from human demonstration and later predicts future robot actions using visual, robot, and haptic information.

Our contribution lies in the following three-fold: 1) We introduce Haptic-Informed ACT, an enhancement of ACT that incorporates haptic feedback to improve grasp failure detection and adaptive correction in contact-rich manipulation tasks.
Unlike the standard ACT, which relies solely on visual and proprioceptive information, Haptic-Informed ACT enables the robot to autonomously detect and retry failed grasps, significantly improving task success rates in dynamic environments. 
We demonstrate that Haptic-Informed ACT effectively improves task robustness and adaptability by leveraging haptic feedback for real-time failure recovery.
Experimental results in pseudo oocyte transfer show that this integration leads to a notable increase in success rates compared to ACT without haptic information.
2) We use failure examples and their recovery behavior in our training data together with successful expert demonstrations to have a recovery-informed training.
This helps the model to adapt to the variations in the environment and especially in cases of failure. 
The robot detects the failure and attempts to recover and retry the task automatically.
3) We developed a novel gripper design made of 3D TPU soft material.
Its flexibility allows the robot to grasp soft experimental objects without damaging their soft tissue. 
This design further enhances the reliability and accuracy of the robot when handling flexible, delicate, and fragile objects while increasing the effectiveness of haptic-based feedback.

The remainder of this paper is organized as follows.
Section~\ref{sec:related_works} introduces previous works on cell manipulation and imitation learning.
Section~\ref{sec:proposed_method} details our proposed system.
Section~\ref{sec:experiments} presents our experimental setup and results.
Finally, Section~\ref{sec:conclusion} concludes this paper with directions for future work.

\section{Related Works}
\label{sec:related_works}
Cell manipulation is fundamental in many applications of biological and biomedical engineering. 
Such as studies of how the cells react to different environments to vitrification of oocytes in fertility treatments.
In these biomedical tasks, high accuracy and carefulness are required when handling oocytes, as they are fragile.
Many related works automate the task of picking and placing oocytes by manipulating them with robots or contactless methods~\cite{zhang2016robotic, zhu2017study, miao2021development}.
Most automation of oocyte manipulation focus on the design of custom hardware for the task or visual algorithms for detecting a target oocyte in a group before its manipulation.
However, a human operator must still select and confirm the target oocyte or manipulate the robot to execute the pick and place task.
We propose training a model to automatically pick and place the target using the advantages of imitation learning and transformers.
To achieve this, the robot needs to be taught how to execute the task by human example by using IL.

Behavior cloning (BC)~\cite{pomerleau1988alvinn, du2022play, mandlekar2021matters} is a type of IL that performs supervised learning from observations to actions; it learns $\pi(a|o)$ where $a$ is the action, and $o$ is the observation.
BC development has been diverse, including approaches that aim to improve the overall performance of not only learning data from one task but also using data from multiple tasks, large-scale data, including natural language, and large-scale models~\cite{brohan2023rt, brohan2022rt, fu2024mobile}.

Although BC has been studied extensively, the success rate is low for tasks requiring delicate and complex manipulations because the robot deviates from the target position; hence, manipulating oocytes is very challenging for normal IL approaches. 
One of the causes of the deviation from the target is the compound error~\cite{zhao_learning_2023, chi2023diffusion, ke2021grasping}, the error from the previous time step accumulates and causes the robot to deviate from the training distribution, making it difficult to return to its proper trajectory.
Zhao~\textit{et~al.} proposed Action Chunking by Transformers (ACT) a low-cost teleoperation environment and imitation learning model for fine-grained tasks~\cite{zhao_learning_2023}.
This method has shown that it can perform appropriate actions for multiple tasks requiring fine manipulation, handle flexible objects, and transfer the location of task-related tasks.

Additionally, multiple studies have demonstrated that ACT can be used with force information to execute fine movements~\cite{kamijo2024learning, kobayashi2024alpha, buamanee2024bi, zhao2024aloha}, such as suturing or inserting Velcro.
It can also be applied to more complex and precise tasks, such as medical suturing, where sub-millimeter accuracy is required~\cite{tanwani2020motion2vec}.

In learning from demonstrations such as IL, the quality of the expert data used to train is very important, normally these samples are always successful cases. 
The logic is that a robot should learn only from good samples and not failures, however, learning from failure or errors ~\cite{grollman2011donut, wong2022error} was explored before. 
In our work, we proposed to use recovery-informed training, by having data samples of the robot failing and its recovery behavior so that during the inference of the model, the robot can adapt and overcome failures that will be included in its training.

\section{Proposed Method}
\label{sec:proposed_method}

This paper presents Haptic-Informed ACT, an advanced multimodal learning framework that builds upon ACT by incorporating haptic feedback, enabling robust and adaptive xenopus oocyte manipulation in dynamic environments.
We use ACT~\cite{zhao_learning_2023}, as inspiration and base for our model.
Action chunking is based on an implementation inspired by the psychological concept of chunking actions as a policy compression~\cite{lai2022action}.
ACT successfully captures the features observed from images by predicting future action sequences and taking advantage of their ensemble and transformers~\cite{vaswani2017attention}, reducing the compounding error that accumulates during sequential decision-making. 

Haptic information is very important in manipulation tasks, to determine if an item is being picked properly. 
Previous works using force information together with ACT~\cite{buamanee2024bi, kamijo2024learning} use the whole torque information of the robot joints.
However, we consider only the haptic information in our region of interest which is the end effector gripper of the robot, by using a 3-axis force sensor, to detect when the gripper is holding something and how much force is being used.

Based on this, Haptic-Informed ACT can predict future action sequences based on previous images, haptic, and action information.
Fig.~\ref{fig:proposed_method} shows the architecture of Haptic-Informed ACT, which models the policy $\pi (a_{t:t+k}|o_t)$, where $a$ are the actions, and $o$ the visual, haptic and robot observations.
Where $a_t^{pos}$, $o_t^{force}$ and $o_t^{img}$ are the observations of position, force, and image respectively.
First, it models the non-Markovian behavior of human demonstrations by action chunking.
Second, it uses a Conditional Variational AutoEncoder (CVAE) to capture the variability in the human demonstration data collected.
Third, the Temporal Ensembling is introduced to improve the smoothness of the measures further.
Instead of predicting actions one step at a time, it predicts the target joint position or action sequence for the next $k$ time steps ($T_p^a$ in Fig.~\ref{fig:proposed_method}).

\begin{figure}[t]
    \centering
   \frame{\includegraphics[width=0.99\linewidth]{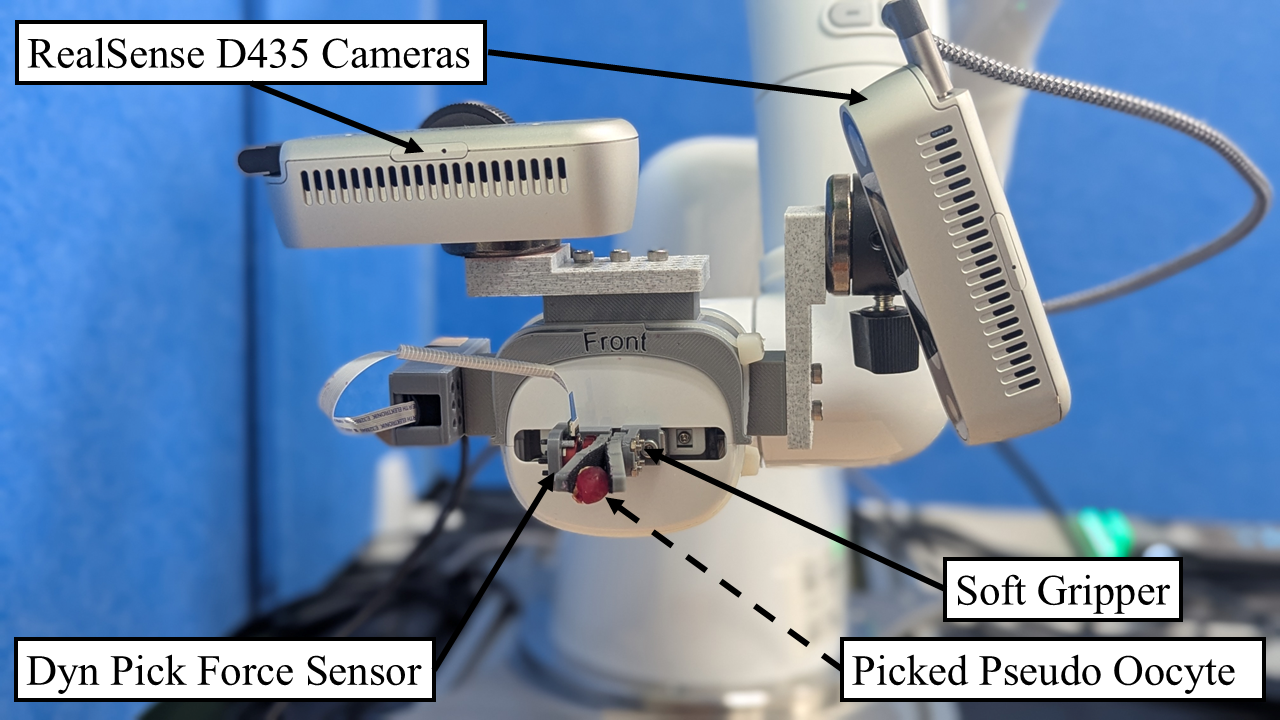}}
    \caption{Cobotta's end effector with the 3D-designed gripper grasping a pomegranate seed as pseudo xenopus oocyte,  two subjective cameras, and the haptic force sensor.}
    \label{fig:robot_ee_soft_gripper}
\end{figure}

Furthermore, we propose to modify the robot's end effector to use a 3D-printed TPU soft material gripper so it does not compress the target pseudo oocyte too much and can adapt to their size variations.
Using a hard gripper it's very challenging to grasp a delicate material without breaking it.
In preliminary experiments, we determined that using hard materials, such as PLA or aluminum, for the gripper fingers will destroy the pseudo xenopus oocytes because they lack adaptability.
To successfully pick a pseudo oocyte with a hard gripper a fine control of the gripper aperture is required, which unnecessarily highly increases the challenge of this task.
To avoid this, we proposed to use a soft gripper, that even when the robot fully closes its gripper, the fingers will bend outwards with the target in the middle, avoiding crushing it, while still having a good grasp of it, as shown in Fig.~\ref{fig:robot_ee_soft_gripper}.
These are the dimensions of the new soft material fingers: 40 mm long, 10 mm wide, and 3 mm of thickness and 3D printed them in flexible TPU material.

\section{Experiments}
\label{sec:experiments}
We propose a biomedical task called the pseudo Xenopus oocyte task, where a pomegranate seed is taken from a cultivation dish and delivered to one of four possible target test tubes.
The pomegranate seed resembles Xenopus oocytes in terms of shape, deformability, and fragility, which is why "pseudo" is included in the task name.

\subsection{Experimental Setup}
\begin{figure}[t]
    \centering
    \frame{\includegraphics[width=0.99\linewidth]{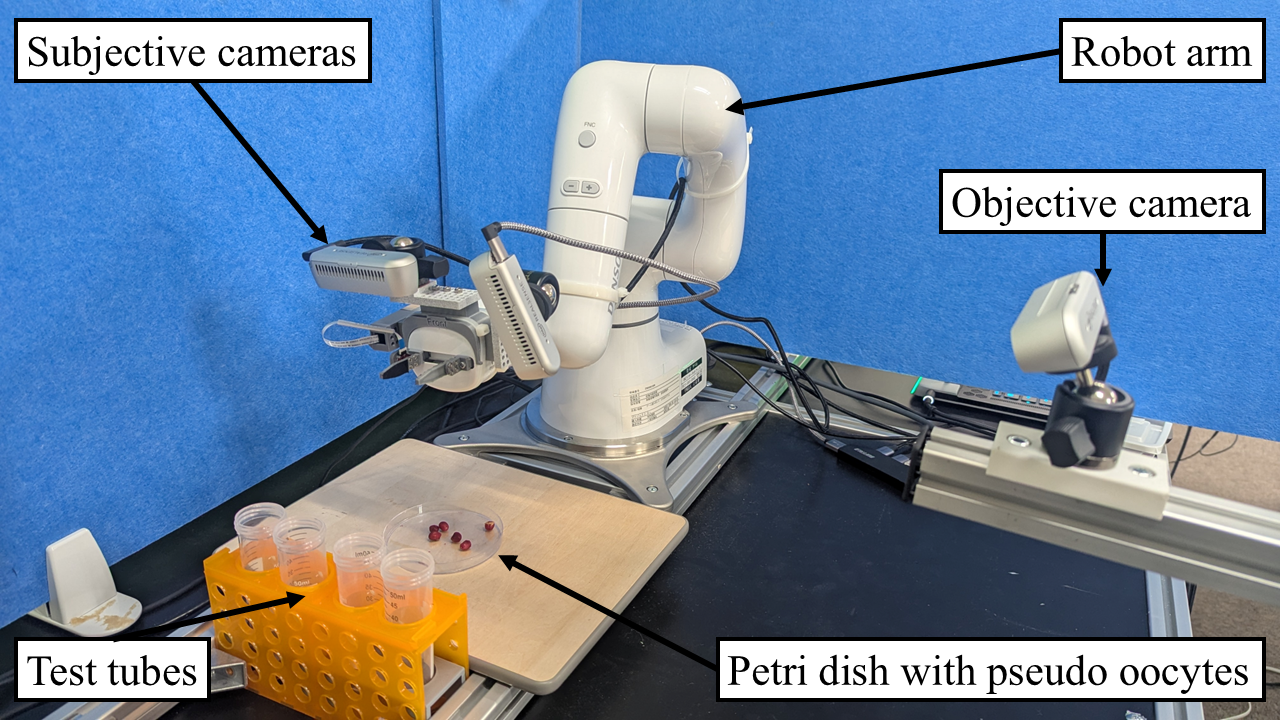}}
    \caption{Robot environment for the pseudo oocyte transfer task.}
    \label{fig:robot_environment}
\end{figure}

The robot arm used is Denso’s Cobotta.
To provide real time feedback on the environment, we use three RealSense D435 cameras: two are attached to the robot end-effector for subjective views, and one provides an objective side view of the environment, as shown in Fig~\ref{fig:robot_environment}.
For haptic feedback, a WACOH Dyn Pick MCF-3 3-axis force sensor is installed.
The robot is controlled using ROS Noetic and MoveIt, and expert demonstration data is collected via teleoperation using a 3D Systems Touch-X haptic device, totaling 50 demonstrations. Of these, 40 are successful demonstrations (including 4 possible target positions for the test tubes), and 10 are recovery demonstrations (where the robot fails to pick up the seed on the first try and needs to retry).
Each trial lasts approximately 25 seconds. The computer used for training is configured with an Intel i9-12900H processor, capable of reaching 5.00 GHz, 32 GB of RAM, and a 16 GB GeForce RTX 3080 GPU, with a total training time of 5 hours.
We compare the proposed Haptic-Informed ACT method with the ACT method, both trained on the same dataset.
To examine the impact of recovery demonstrations, we also conduct ablation experiments with and without these samples.
In the testing phase, we design the following two types of experiments.

\subsection{Trained Environments Tests}
We first execute 10 trials of the pseudo Xenopus oocyte transfer task in the same environment as the training setup and then compare the success rates of different methods.
Each experiment involves four target test tubes, and the number of seeds in the petri dish is randomized (between 1 and 7 seeds).
The petri dish is roughly positioned at the center of the table, with a variation range of $\pm$5 cm along the x and y axes.

\begin{figure*}[t]
    \centering
    \includegraphics[width=0.90\linewidth]{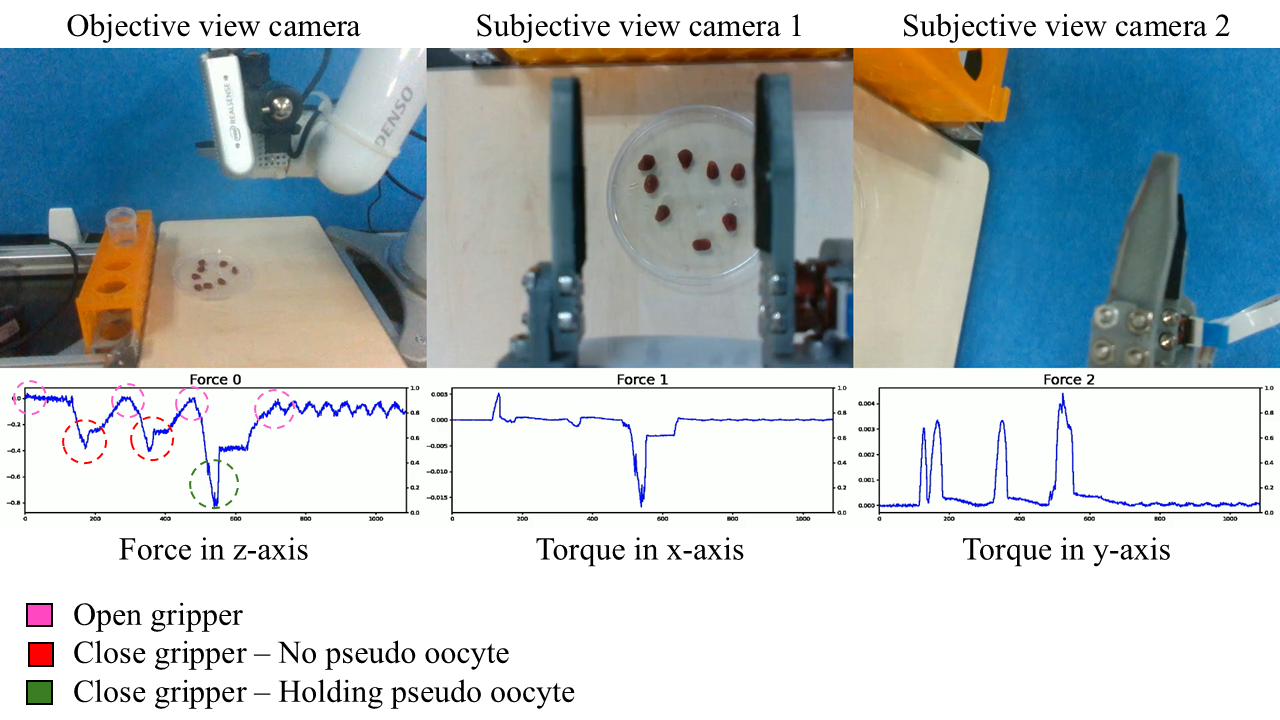}
    \caption{Visual and haptic information of the robot during an experiment. The force in the z-axis shows the different states of the gripper during the task, and how the force differs when the gripper is closed with or without the pseudo oocyte.}
    \label{fig:force_results}
\end{figure*}

The experimental results can be seen in Table~\ref{tab:results_experiments}.
We observe that in all cases, Haptic-Informed ACT outperforms ACT, with a success rate that is twice as high. Specifically, the models trained with recovery behavior data perform better, regardless of whether they are Haptic-Informed ACT or ACT.
In these experiments, when the robot fails to pick the seed from the petri dish, it continues attempting to pick it until successful, and only then moves to the test tube. This behavior was not observed in the trials of ACT with recovery data, which supports our claim that visual information alone is insufficient to determine whether the robot has successfully picked the seed.
This proves that haptic feedback is essential for executing fine manipulation tasks.
Fig.~\ref{fig:force_results} shows both subjective and objective views of the camera and the changes in the haptic sensor during the experiment, highlighting the differences in the gripper’s state when closing with or without the seed.

During the trials, the ACT model simply attempted to pick the seed from the petri dish and, regardless of success or failure, proceeded to the test tube for delivery after each attempt.
In contrast, Haptic-Informed ACT continued attempting to pick the seed until successful.
Haptic-Informed ACT failures occurred when the robot tried to pick the seed but failed, getting stuck in a loop of repeated attempts until the trial time ran out.
In the training data, 20\% of the samples were recovery behavior data.
If this percentage were too high, it could negatively affect the training and performance of the model.
In the future, we plan to further investigate how recovery behavior samples influence the model’s training.
Furthermore, throughout all the experiments, the robot never crushed any seeds, which validates the advantage of using soft gripper materials—regardless of whether the robot picked the seed from the edge, the seed was delivered without any damage.
\begin{table}[t]
    \caption{ Success rate of the Experiments between Haptic-Informed ACT and ACT in a known environment
    }{
    \begin{tabularx}{1.0\linewidth}{|C|C|C|}
        \hline
        {Method} & {Recovery samples} & {w/o Recovery samples}  \\ 
        \hline
        {ACT} & {50\%} & {20\%}  \\
        \hline
        {Haptic-Informed ACT} & {\textbf{80\%}} & {\textbf{40\%}} \\
        \hline
    \end{tabularx}
    }
    \label{tab:results_experiments}
\end{table}

\subsection{Unknown Environment Tests}
\begin{figure}[t]
    \centering
    \frame{\includegraphics[width=0.90\linewidth]{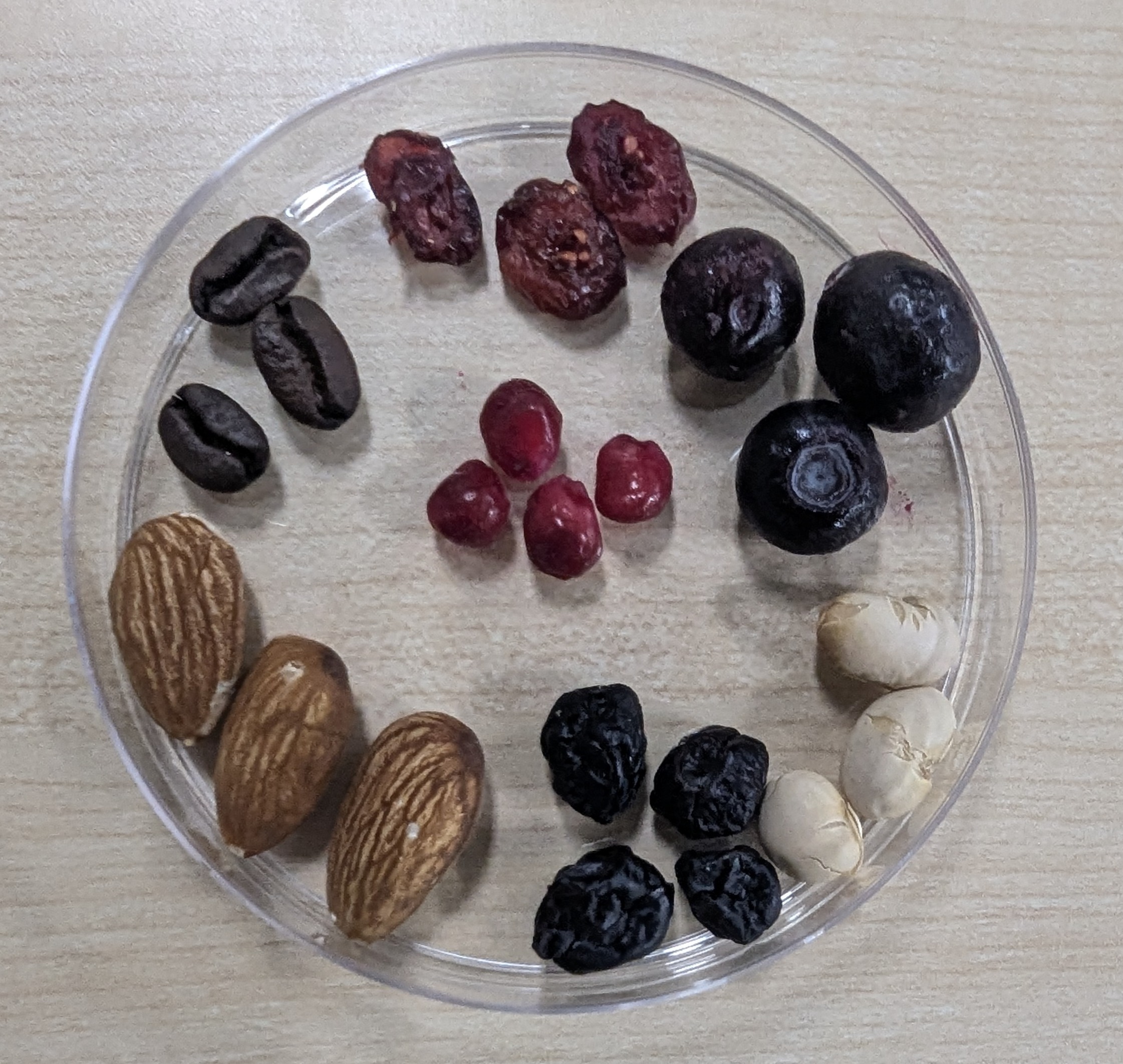}}
    \caption{Pseudo oocytes used in the experiments. Clockwise from the top: Dried cranberries, frozen blueberries, fried soybeans, dried blueberries, almonds, coffee beans, and pomegranate seeds. }
    \label{fig:target_objects}
\end{figure}

\begin{table}[t]
\caption{Different dimensions of the pseudo xenopus oocytes}
\begin{center}
    \label{tab:materials}
    \begin{tabularx}{0.98\linewidth}{|L|c|c|c|}
    \hline
    \textbf{Pseudo oocyte} & \textbf{Shape} & \textbf{Color} & \textbf{Size LxWxH [mm]}\\
    \hline
    Pomegranate seed & Ellipsoid & Red & 8.9 x 5.42 x 5.42 \\
    \hdashline
    Dried blueberry & Sphere & Dark blue & 10.31 x 9.73 x 7.68\\
    \hdashline
    Frozen blueberry & Sphere & Dark blue & 10.68 x 13.26 x 13.26\\
    \hdashline
    Dried cranberry & Ellipsoid & Red & 13.40 x 9.56 x 5.77\\
    \hdashline
    Fried Soybean & Sphere & Beige & 13.86 x 8.38 x 8.65\\
    \hdashline
    Coffee bean & Ellipsoid & Dark brown & 14.42 x 9.03 x 5.40\\
    \hdashline
    Almond & Ellipsoid & Dark brown & 23.38 x 12.70 x 7.78\\
    \hline
    \end{tabularx}
\end{center}
\end{table}

To evaluate the model’s generalization capabilities, we tested it in an unseen environment, including the use of entirely new pseudo oocytes and changes in the background.
Specifically, we simulated different Xenopus oocytes using materials that differ in size, color, and shape from the pomegranate seed, including dried blueberries, frozen blueberries, dried cranberries, fried soybeans, coffee beans, and almonds, as shown in Fig.~\ref{fig:target_objects}, with their detailed dimensions listed in Table~\ref{tab:materials}. 
We executed 10 trials for each material and evaluated their success rates in picking and delivering the seed.
The experimental results can be seen in Table~\ref{tab:results_unknown}. 
We considered the success rate of picking the pseudo oocyte and delivering it to the test tube separately, as there were cases where the robot could pick the seed but failed during the delivery. 
In all experiments, the robot successfully picked and delivered the seeds, with the worst case occurring when using almonds as a target.
The failure was due to the almond’s shape; when the long side of the seed was parallel to the gripper, it was easy to pick, but when the gripper closed, one of the robot’s fingers would push the almond outside the gripper’s range.
In the dried blueberry trials, the berries began to melt and slipped out of the gripper, getting damaged when they fell onto the table. 
Additionally, the melting blueberries caused red juice drops to stain the petri dish, which affected the model’s performance by confusing it into mistaking the red drops as targets to pick.

In these experiments, we observed the model’s ability to handle covariance shift, a capability inherited from the original ACT architecture. The difference in target color did not affect the robot’s ability to locate the target, especially in the most challenging case, the fried soybean, whose color closely resembled that of the table.
An interesting behavior was observed in the almond and frozen blueberry trials: even when the robot successfully picked the target, the force detected on the gripper, due to the size difference, was greater than the force experienced during training with the pomegranate seed.
The robot then misinterpreted this as a failure and attempted to pick again, eventually getting stuck in a loop.
 
Finally, in all of these experiments, the soft gripper was able to pick the items, regardless of size differences, without breaking or damaging them

\begin{table}[t]
\caption{Results of the experiments with new pseudo xenopus oocytes}
\label{tab:results_unknown}
\begin{center}
    \begin{tabular}{|c|c|c|c|c|}
    \hline
    Pseudo oocyte & Success rate pick & Success rate delivery\\
    \hline
    Dried blueberry & 60\% & 60\% \\
    Frozen blueberry & 50\% & 40\% \\
    Dried cranberry & 50\% & 50\% \\
    Soybean & 60\% & 60\% \\
    Coffee bean & 40\% & 40\% \\
    Almond & 20\%& 20\% \\
    \hline
    \end{tabular}
\end{center}
\end{table}

\section{Conclusion}
\label{sec:conclusion}

In this paper, we proposed Haptic-Informed ACT which uses haptic information together with action chucking with transformers to automate a pseudo oocyte transfer task.
Haptic-Informed ACT proved to be able to successfully execute the task outperforming ACT which was used as the baseline.
Furthermore, our model can adapt to changes in the environment, such as variations in the number, color, size, and position of oocytes and daily changes in lighting conditions.
Additionally, we proposed to use a 3D-printed soft gripper to handle the pseudo oocytes without damaging them, this proved to be useful when the size, and shape of the targets changed.

In future work, we will consider improving Haptic-Informed ACT's by further exploring how the introduction of recovery behaviors in the recovery-informed training affects the performance of the model, by doing ablation studies.
We are also considering the possibility of predicting future observations and action sequences from past observations and action sequences.
This is based on the idea of predicting future senses together with the original Haptic-Informed ACT.

\section*{Acknowledgment}
This work was supported by the Japan Science and Technology Agency (JST) Moonshot Research \& Development Program, Grant Number JPMJMS2033.


\balance


\bibliographystyle{templates/IEEEtran}
\bibliography{references.bib}


\end{document}